\documentclass[letterpaper, 10 pt, conference]{ieeeconf}
\IEEEoverridecommandlockouts
\usepackage{cite}
\usepackage{amsmath,amssymb,amsfonts}
\usepackage{algorithmic}
\usepackage{hyperref}
\usepackage{graphicx}
\usepackage{textcomp}
\usepackage{xcolor}

\usepackage[english]{babel}
\usepackage[T1]{fontenc}

\usepackage[nolist]{acronym}
\usepackage{amsmath}
\usepackage[font={small}]{caption}
\usepackage{subcaption}
\usepackage[export]{adjustbox}
\usepackage{amssymb}
\usepackage{bm}
\usepackage{booktabs}
\usepackage{booktabs}
\usepackage{balance}
\usepackage{cite}
\usepackage{color}
\usepackage{dsfont}
\usepackage{esvect}
\usepackage{float}
\usepackage{graphicx}
\usepackage{hyperref}
\usepackage{import}
\usepackage{mathtools}
\usepackage{multirow}
\usepackage{multicol}
\usepackage{multirow}
\usepackage{flushend}
\usepackage{outline}
\usepackage{paralist}
\usepackage{siunitx}
\usepackage{stackengine}
\usepackage{stmaryrd}
\usepackage{systeme}
\usepackage{soul}
\usepackage{url}
\usepackage{tikz}
\usetikzlibrary{tikzmark}
\usetikzlibrary{calc}
\usepackage[normalem]{ulem}

\definecolor{sbrown}{rgb}{0.7254, 0.2784, 0.0} 
\definecolor{syellow}{rgb}{0.9412, 0.7020, 0.1373} 
\definecolor{syellowdark}{rgb}{0.7059, 0.5265, 0.1030} 
\definecolor{sgrey}{rgb}{0.5373, 0.5529, 0.5529} 
\definecolor{sgreydark}{rgb}{0.4030, 0.4147, 0.4147} 
\definecolor{sblue}{rgb}{0.0, 0.6118, 0.8706} 
\definecolor{sbluedark}{rgb}{0.0, 0.4589, 0.6530} 
\definecolor{sgreen}{rgb}{0.2314, 0.6157, 0.5529} 
\definecolor{sgreendark}{rgb}{0.1764, 0.4588, 0.4118}

\newcommand{\projectwebsite}{https://tif-twirl-13.github.io/learning-compliance}

\newcounter{nodecount}
\newcommand\tabnode[1]{\addtocounter{nodecount}{1} \tikz \node (\arabic{nodecount}) {#1};}

\tikzset{every picture/.append style={remember picture,baseline},
         every node/.append style={inner sep=0pt,anchor=base,
         text depth=.25ex,outer sep=1.5pt},
         every path/.append style={thick, dashed, rounded corners}}

\begin{document}

\title{\LARGE \bf 
Learning Force Control for Legged Manipulation
}

\author{
Tifanny Portela$^{12}$, Gabriel B. Margolis$^1$, Yandong Ji$^{13}$, and Pulkit Agrawal$^1$
\thanks{Research was conducted in the Improbable AI Lab at MIT. Author affiliations: $^1$ Improbable AI Lab. $^2$ EPFL. $^3$ University of California, San Diego.}
}

\let\oldtwocolumn\twocolumn
\renewcommand\twocolumn[1][]{%
    \oldtwocolumn[{#1}{
    \begin{flushleft}
   \centering
    
    \begin{minipage}{0.1\textwidth}
        \centering
        \textit{Compliance}
    \end{minipage}%
    \begin{minipage}{0.8\textwidth}
        \centering
        \includegraphics[width=0.22\textwidth]{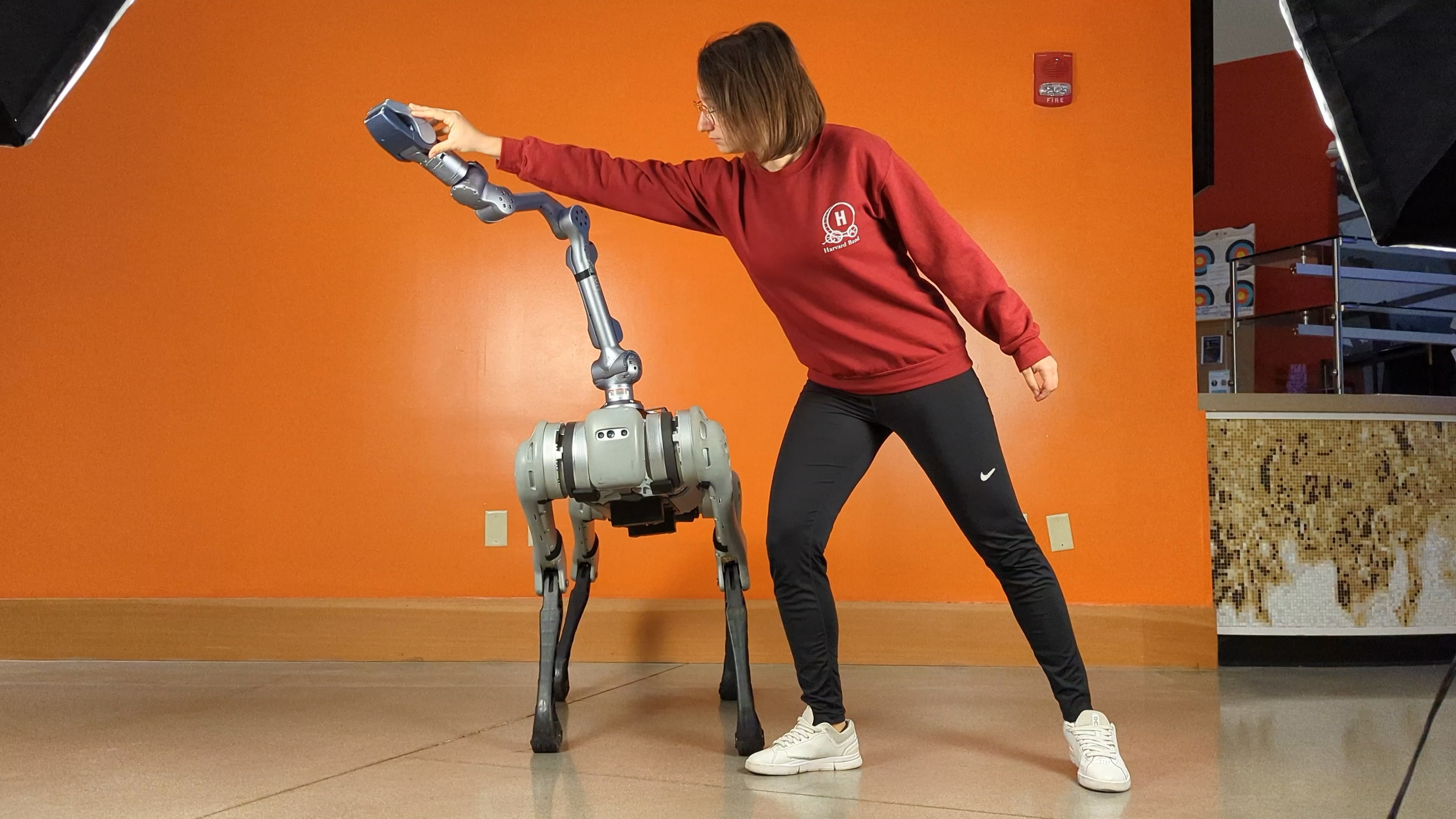}
        \includegraphics[width=0.22\textwidth]{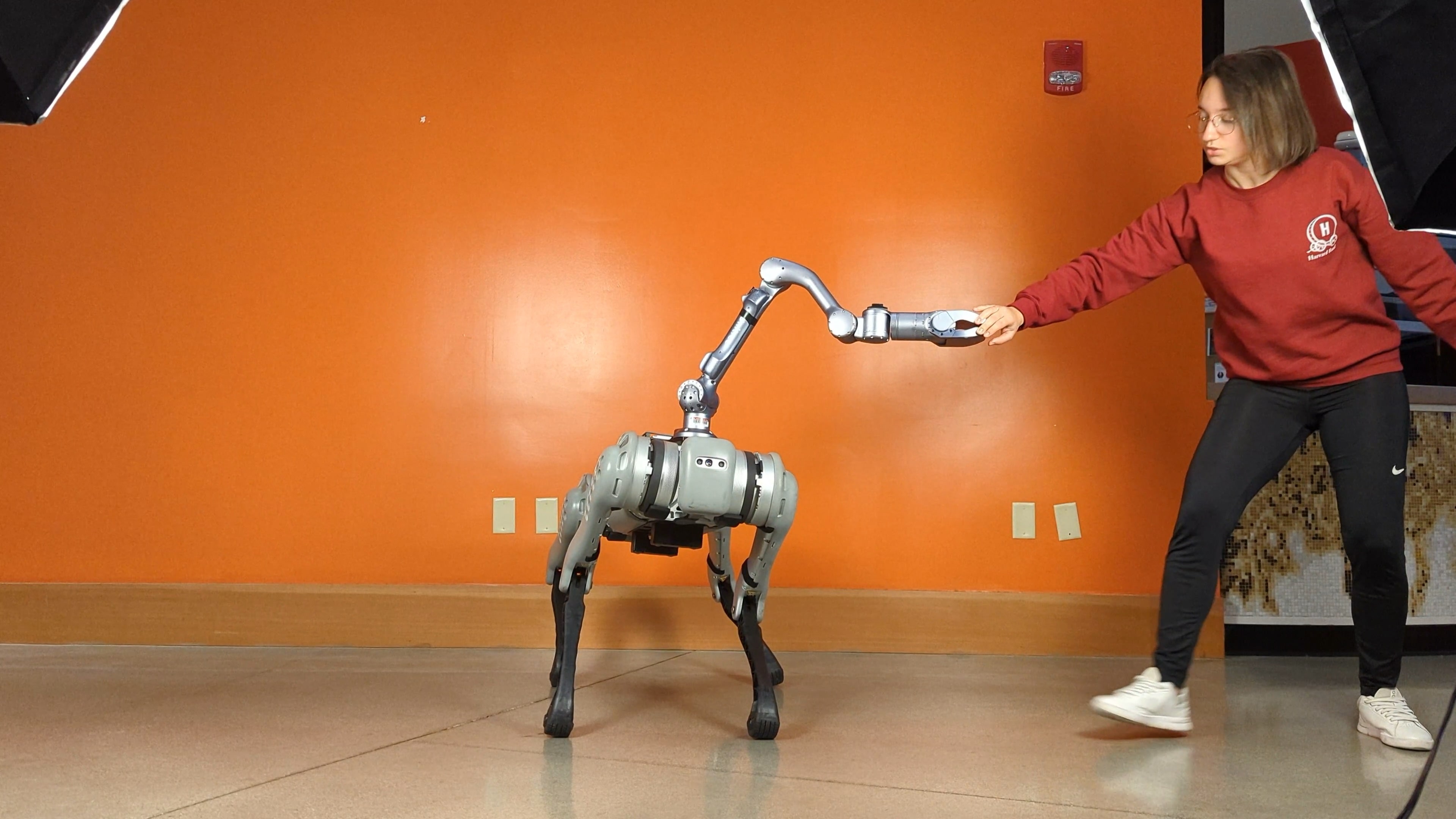}
        \includegraphics[width=0.22\textwidth]{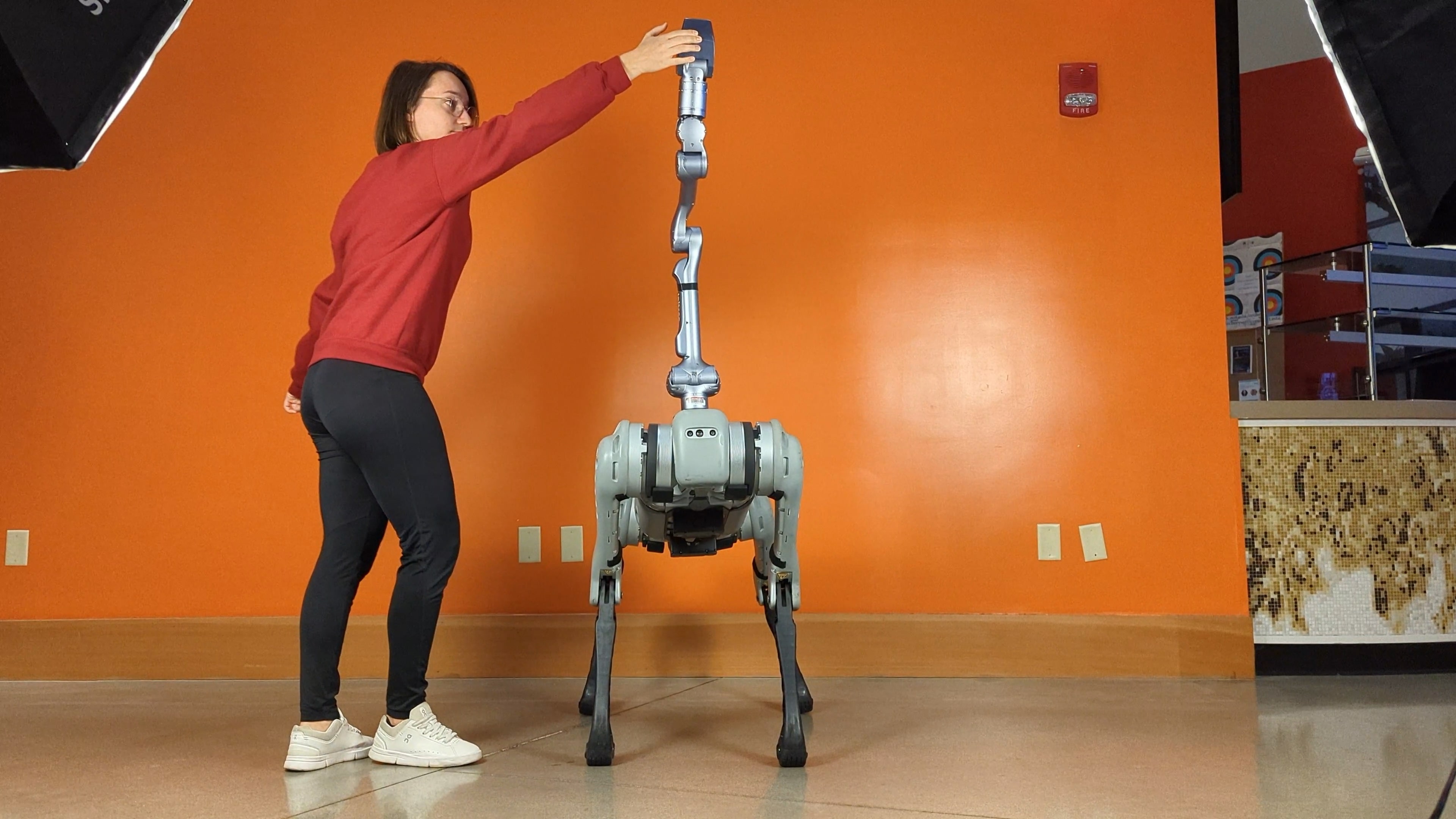}
        \includegraphics[width=0.22\textwidth]{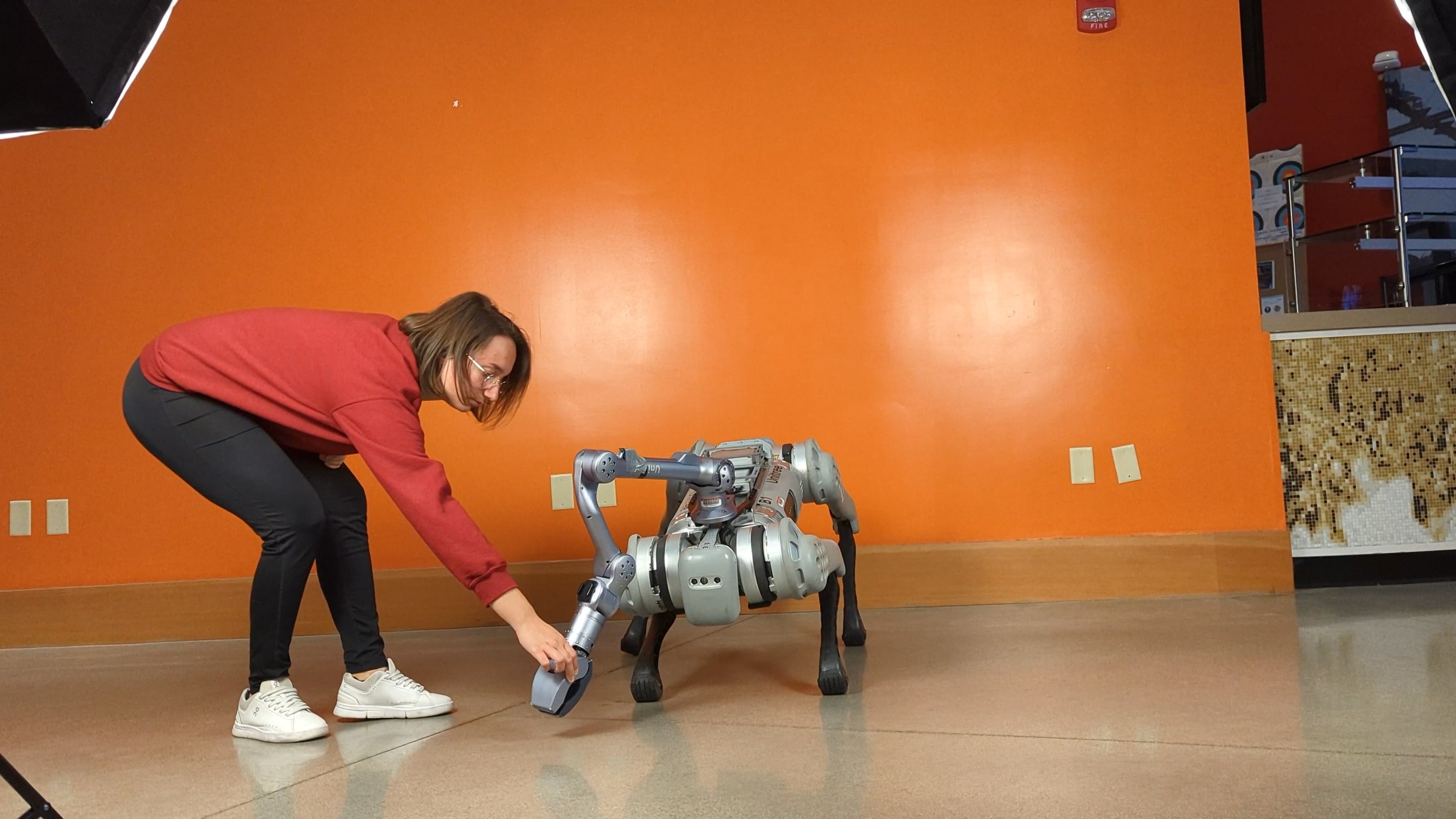}
    \end{minipage}
    
    \vspace{0.2cm}

    \begin{minipage}{0.8\textwidth}
        \centering
        \includegraphics[width=0.22\textwidth]{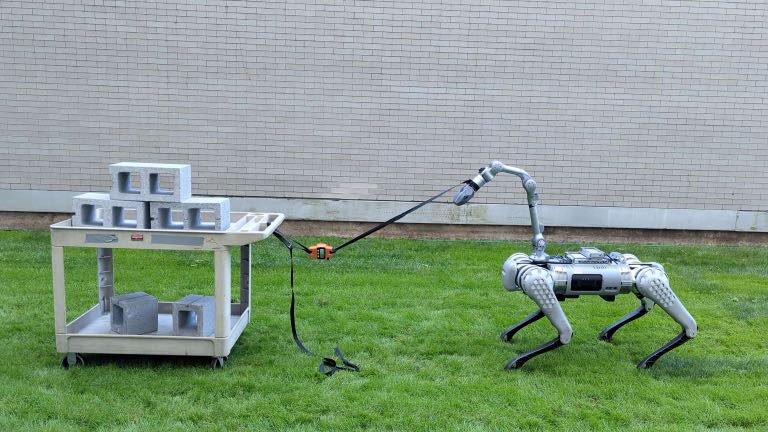}
        \includegraphics[width=0.22\textwidth]{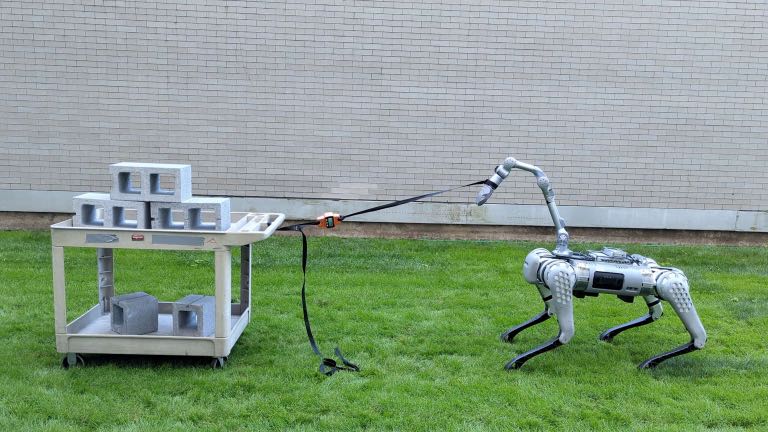}
        \includegraphics[width=0.22\textwidth]{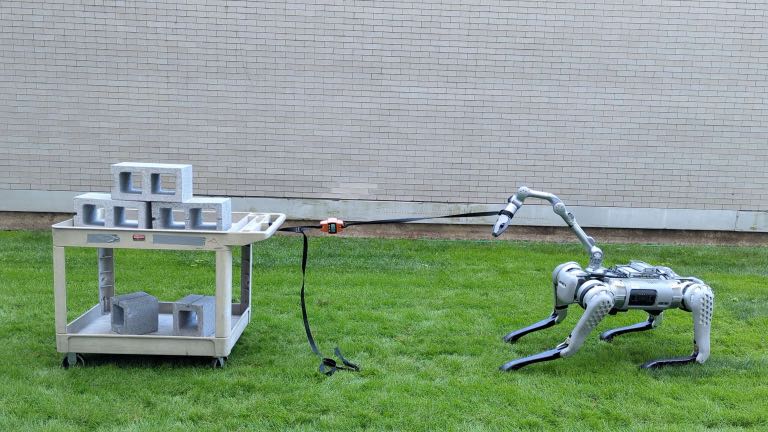}
        \includegraphics[width=0.22\textwidth]{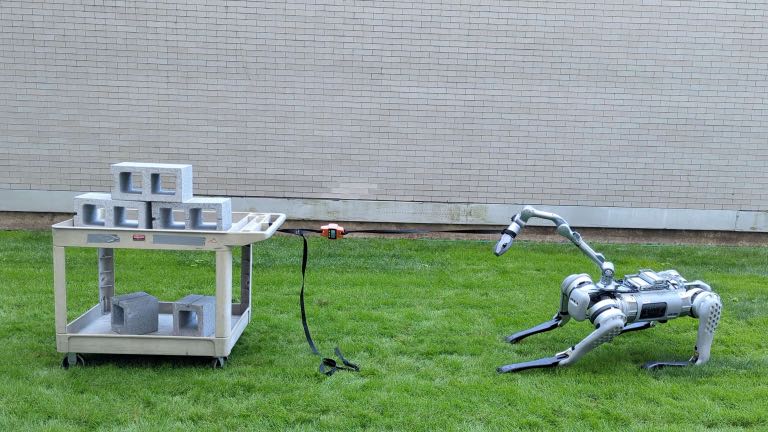}
    \end{minipage}
    \begin{minipage}{0.1\textwidth}
        \centering
        \textit{Whole-body Pulling}
    \end{minipage}%

    \vspace{0.2cm}

    \begin{minipage}{0.1\textwidth}
        \centering
        \textit{Loco-Manipulation}
    \end{minipage}%
    \begin{minipage}{0.8\textwidth}
        \centering
        \includegraphics[clip,trim=0cm 0cm 0cm 0cm,width=0.22\textwidth]{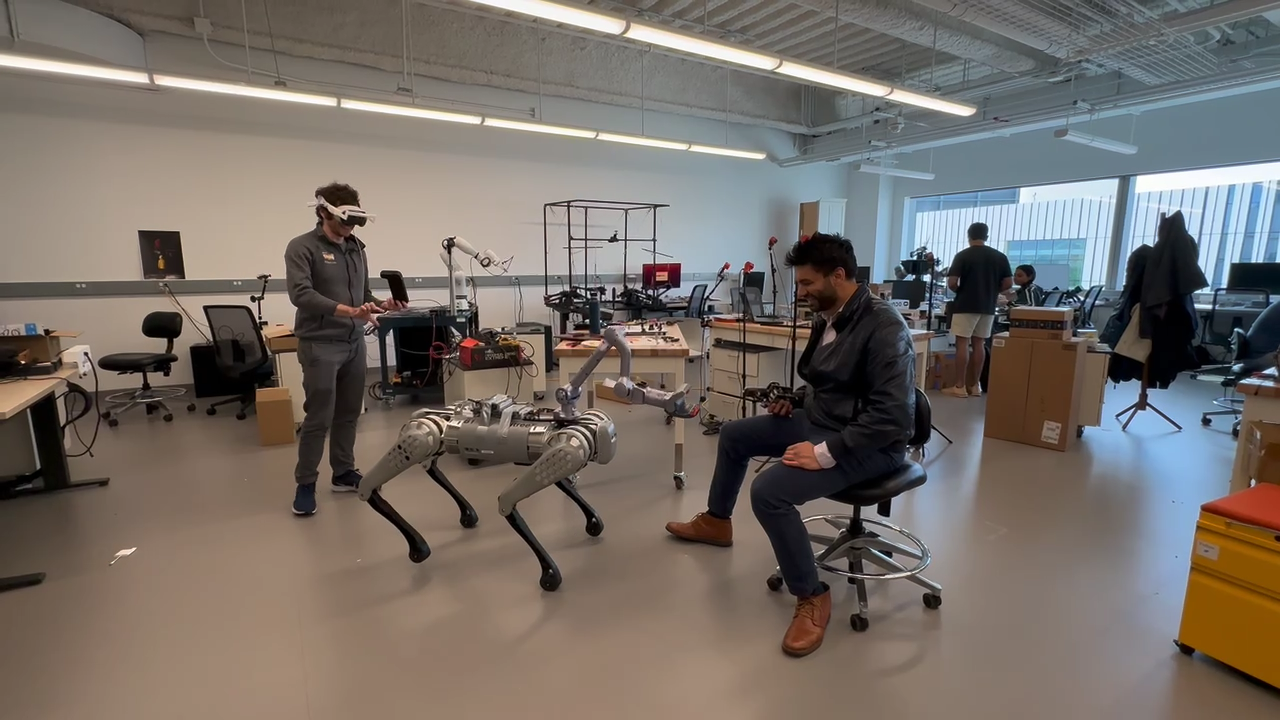}
        \includegraphics[clip,trim=0cm 0.07cm 0cm 0.1cm,width=0.22\textwidth]{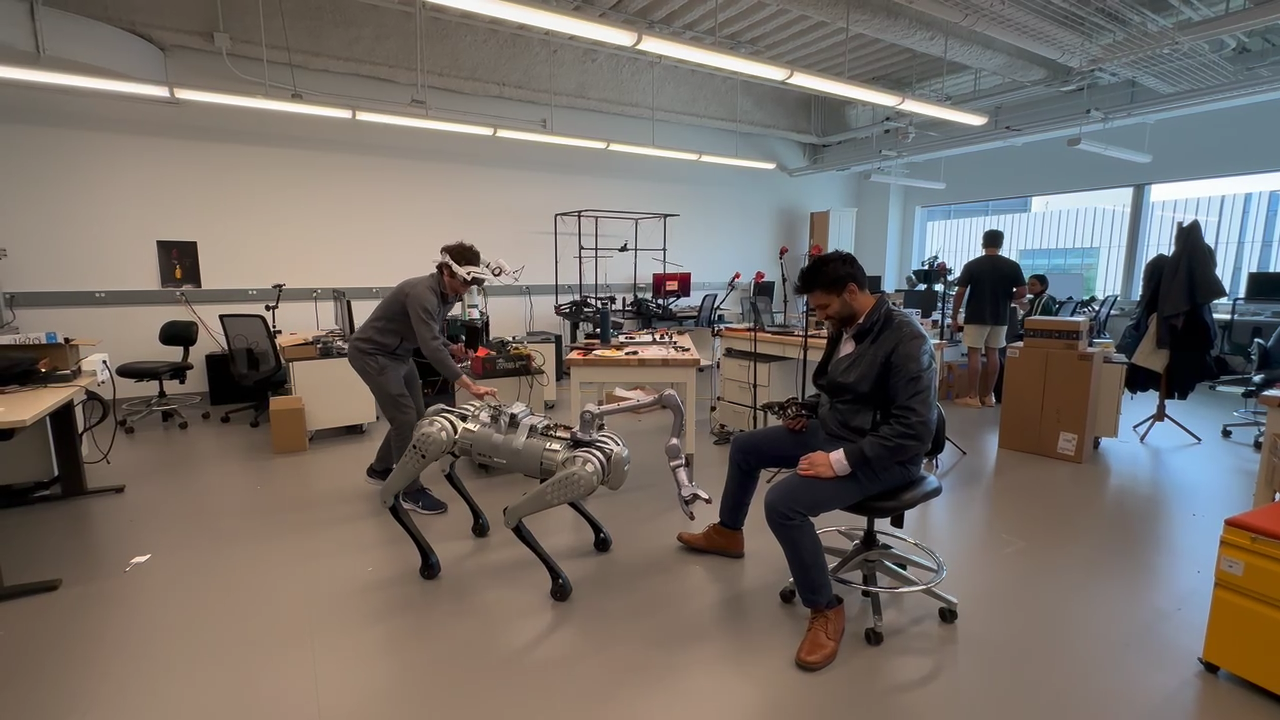}
        \includegraphics[clip,trim=0cm 0.145cm 0cm 0cm,width=0.22\textwidth]{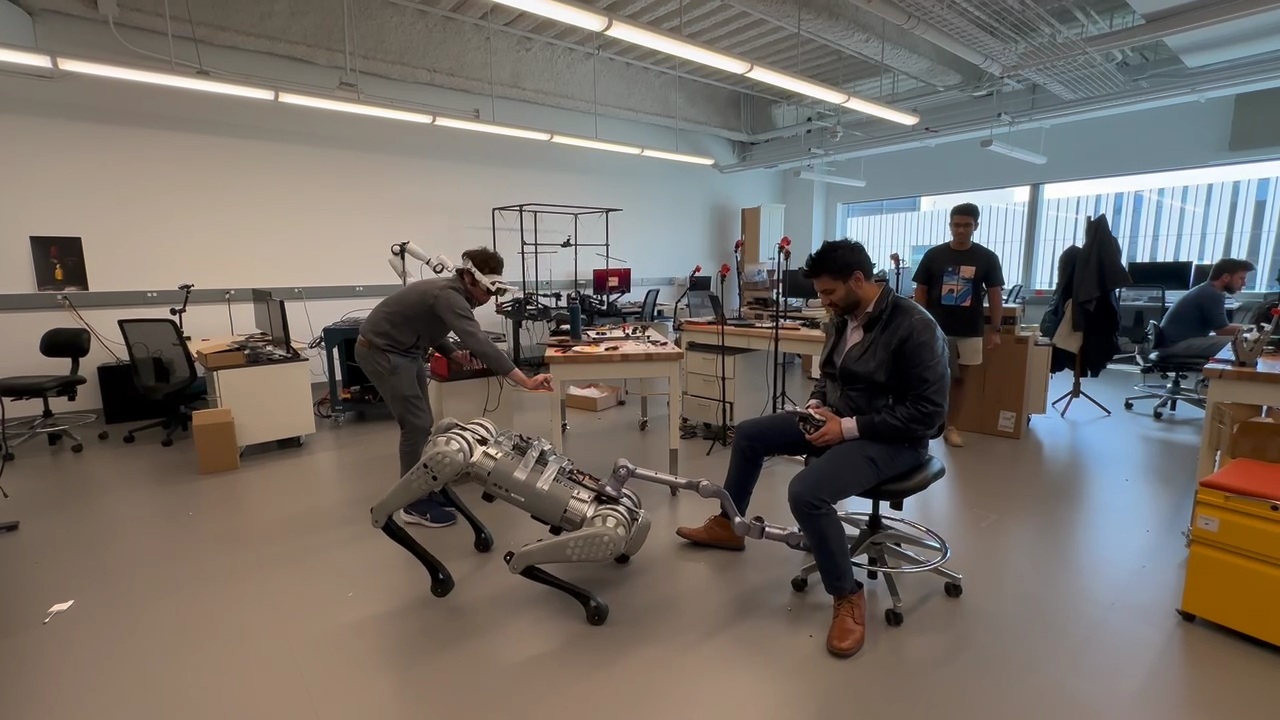}
        \includegraphics[clip,trim=0cm 0.22cm 0cm 0cm,width=0.22\textwidth]{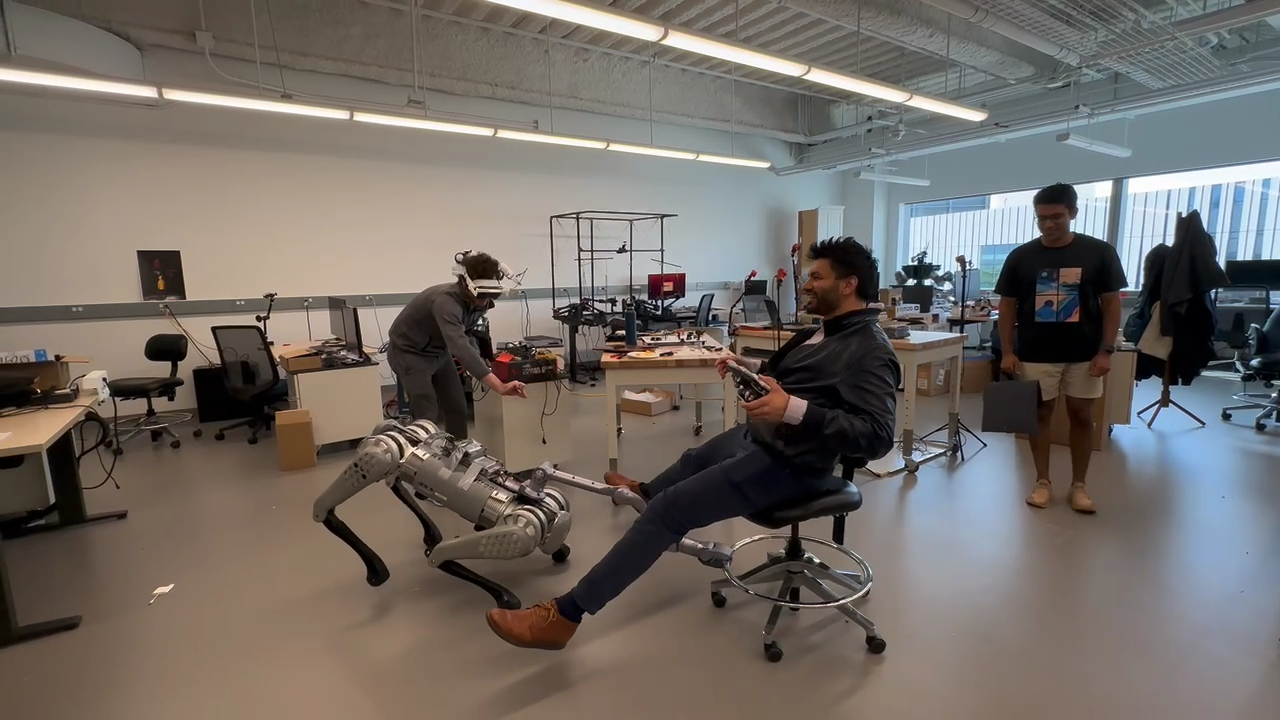}
    \end{minipage}
    
    \vspace{0.2cm}


    \vspace{0.3cm}
    
    \captionof{figure}{
    We train a whole-body policy to control the \textit{force} applied at the end effector of a legged manipulator.
    This enables compliant interaction (top) and whole-body force application (middle). 
    Closing the loop to perform impedance control allows the teleoperator to modulate the gripper's compliance during loco-manipulation (bottom).
    }\label{fig:main}
    \end{flushleft}
    }]
}

\maketitle
\thispagestyle{empty}
\pagestyle{empty}

\begin{abstract}

Controlling contact forces during interactions is critical for locomotion and manipulation tasks. While sim-to-real reinforcement learning (RL) has succeeded in many contact-rich problems, current RL methods achieve forceful interactions implicitly without explicitly regulating forces. We propose a method for training RL policies for direct force control without requiring access to force sensing. We showcase our method on a whole-body control platform of a quadruped robot with an arm. Such force control enables us to perform gravity compensation and impedance control, unlocking compliant whole-body manipulation. The learned whole-body controller with variable compliance makes it intuitive for humans to teleoperate the robot by only commanding the manipulator, and the robot's body adjusts automatically to achieve the desired position and force. Consequently, a human teleoperator can easily demonstrate a wide variety of loco-manipulation tasks. To the best of our knowledge, we provide the first deployment of learned whole-body force control in legged manipulators, paving the way for more versatile and adaptable legged robots. See our robot in action: \href{\projectwebsite}{\texttt{tif-twirl-13.github.io/learning-compliance}}.
\end{abstract}

\section{Introduction}

Legged manipulators offer unique advantages over fixed-base and wheeled manipulators. For example, they can climb stairs, crouch under a sink to perform a repair job, and reach objects inaccessible to wheeled robots by shifting their center of mass to facilitate a larger workspace and stronger posture.
In recent years, reinforcement learning (RL) algorithms have achieved state-of-the-art results in legged locomotion~\cite{hwangbo2019learning, kumar2021rma, rudin2022learning, margolis2022rapid}, dexterous manipulation~\cite{chen2021system,chen2022visual}, and drone flight~\cite{kaufmann2023champion}. These successes have led to an increased interest in applying RL to tasks that require a robot to interact forcefully with its environment. Successfully performing forceful tasks such as running, kicking, parkour, and fall recovery~\cite{fu2023deep, ji2023dribblebot, ma2023learning, hoeller2023anymal, zhuang2023robot, cheng2023extreme} requires implicit regulation of the force between a robot and its surroundings. However, policies learned with RL often lack the expressivity to control forces explicitly, limiting the user's ability to set the robot's compliance to balance safety with forceful interaction.

In this work, we propose to learn policies wherein the user can directly command the policy to exert a specific contact force through a mobile manipulator on a quadruped robot. We characterize the ability of learned policies to control and estimate the specified forces. Our system expands the capability of policies learned using RL in forceful manipulation tasks. Our learned force control is particularly useful in various scenarios, such as enabling compliance for kinesthetic demonstration, safer human-robot interaction, and optimizing the whole-body posture for applying large forces across a larger workspace. We can leverage our force controller to easily perform impedance control wherein the gripper tracks a position setpoint with controllable stiffness.  
 
To demonstrate the capabilities of our whole-body force control on multiple tasks, we teleoperate the robot to perform tasks that emphasize force application across a large workspace. A challenge in teleoperating quadruped manipulators is the morphology mismatch between the robot and the human operator. We address this challenge following prior work that learns a policy that takes the teleoperation command for moving the manipulator as input and outputs joint position commands for each motor of the robot\cite{fu2023deep}. The policy is learned using simulated data and transferred to reality. Such a setup makes it intuitive for a human to control the manipulator -- the human directly commands the manipulator without worrying about commanding all robot joints -- a function performed by the learned policy.  
Our main contribution is the ability to directly command forces, which allows the teleoperator to modulate the compliance of the end effector and, thereby, realize gravity compensation and impedance control. Our method for force control doesn't require access to force-torque sensors. Instead, it relies on proprioceptive sensing to estimate the forces applied by the robot and modulate them as necessary. The method for training  the force controller is described in Section~\ref{sec:method} and Section\ref{sec:results} characterizes the effectiveness of force tracking.

\section{Related Work}

\subsection{Reinforcement Learning for Loco-Manipulation}
Prior work has controlled a quadruped with mounted arm using sim-to-real reinforcement learning by formulating the task as simultaneous end effector position control and locomotion velocity control \cite{fu2023deep, lee2022learning}. Our work is highly influenced by \cite{fu2023deep}, which was deployed on a real robot and displayed an increased workspace. However, it proposed no explicit means to control force or compliance at the end effector. Other works have learned a policy for a downstream task that requires forceful interaction, such as preventing and recovering from falls with a mounted arm \cite{ma2023learning} or dribbling a ball with the feet \cite{ji2023dribblebot, haarnoja2023learning}. These works demonstrate that forceful interaction is possible but optimize it in service of a single task. Without an explicit way of commanding the interaction force, we cannot repurpose these controllers for other force application tasks, and the precision of the force control in simulation and reality cannot be directly evaluated. Another line of work has used hierarchical architecture to push large objects using the robot's body, without an arm \cite{jeon2023learning}. While the body and legs alone can perform useful environment interactions, a mounted arm can increase the workspace and allow the robot to control the direction and magnitude of force and torque application more precisely.

\subsection{Forceful Control Primitives}
For some manipulation tasks, particularly those involving contact interactions, it is hard to teleoperate by commanding position setpoints due to the unknown geometry of the contact surface or a desired degree of compliance in the motion. Instead, various parameterizations of control that regulate contact force have been proposed. Classic work on compliance and force control \cite{mason1981compliance, raibert1981hybrid, yoshikawa1987dynamic, hogan1984impedance} established impedance control and hybrid force-velocity control. The purpose of these techniques is to define how the robot should reconcile force and position tracking errors given that these quantities cannot be independently controlled during contact. Force control methods may be implemented with closed-loop feedback from a wrist force sensor or with less precision by estimating the contact force from proprioceptive actuators and robot model~\cite{chiaverini1999survey}. Our approach does not strictly correspond to any classical control formulation because our objective function combines regularization terms common in reinforcement learning with a force tracking objective.

One use case for force control is kinesthetic teaching, wherein a human physically interacts with a robot to demonstrate the movements and forces it should apply on the environment to accomplish a task~\cite{calinon2007learning, kormushev2011imitation}. To receive a kinesthetic demonstration, it is necessary for the robot to comply with the operator's guidance and also estimate the amount of force exerted. This method of teaching has previously been considered challenging for legged robots and high-dimensional systems because of the cognitive challenge of reasoning about many degrees of freedom~\cite{ravichandar2020recent}, which provides a motivation for developing low-level controllers that assume some control authority and ease the task.

\begin{figure}[t!]
\centering
\vspace{0.2cm}
\includegraphics[width=8.8cm]{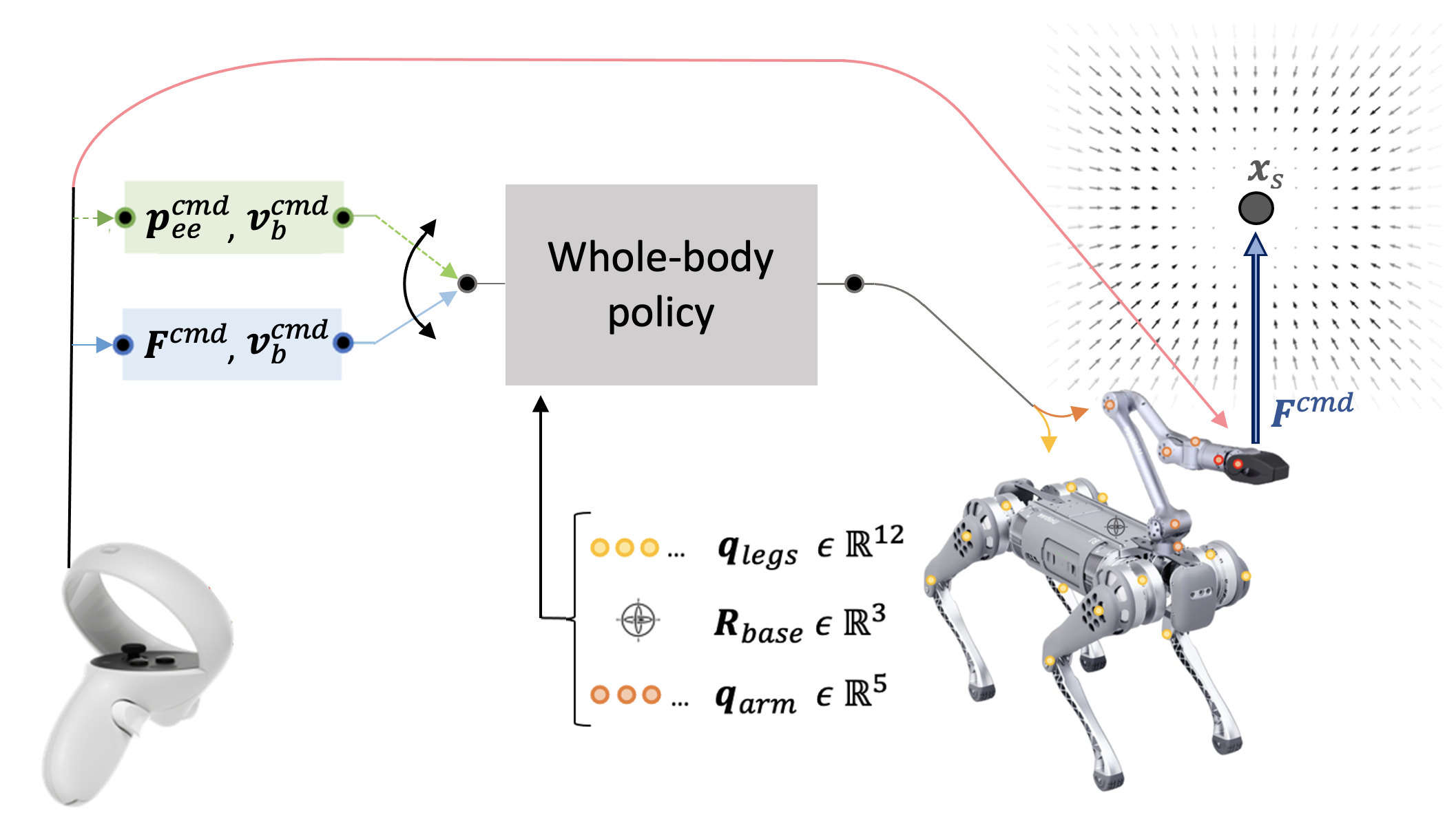}
\caption{\textbf{System architecture.} In addition to the position command $\textbf{p}_{ee}^{cmd}$, we consider training a policy to track a commanded end effector force $\textbf{F}^{cmd}$ while walking following a base velocity command $\textbf{v}_{b}^{cmd}$. A human operator defines the commands using a joystick or hand-tracking headset~\cite{park2024avp}. To learn force tracking, we train the policy in a potential field where the gripper is pulled towards a randomized setpoint $x_s$ (right).}
\label{fig:system_diagram}
\end{figure}

\subsection{Whole-body Control}

Effective model-based approaches have been developed for controlling legged manipulators in a variety of tasks~\cite{dai2014whole, sentis2005synthesis, posa2014direct, sleiman2021unified, polverini2020multi, murphy2012high, murooka2015whole, rehman2016towards, bellicoso2019alma, risiglione2022whole, sleiman2023versatile}. Such works typically optimize a whole-body inverse dynamics model to stabilize a reference trajectory computed using model-predictive control or trajectory optimization. A few works have specialized in forceful interaction with a legged manipulator~\cite{murphy2012high, murooka2015whole, rehman2016towards, bellicoso2019alma, risiglione2022whole}. Early approaches \cite{murphy2012high} used trajectory optimization to generate open-loop behaviors for manipulating heavy objects and a separate online tracking controller to realize the motion. Subsequent work \cite{murooka2015whole} incorporated contact point planning and control to manipulate objects using the entire body of a humanoid robot. More recently, ALMA \cite{bellicoso2019alma} proposed a motion planning and control framework capable of coordinating dynamic and compliant locomotion and mobile manipulation based on a whole-body control task hierarchy. They demonstrated accurate contact force control through the end effector using the whole body. The ALMA system was also shown to support compliant behavior in service of a collaborative payload-carrying task.

\section{Materials}

\textbf{Robot Quadruped}: We used Unitree B1 quadruped robot for all our experiments. This \SI{55}{\kilo\gram} robot stands \SI{0.64}{\meter} tall. It has \SI{12}{} identical electric actuators -- each equipped with a joint position encoder -- and an inertial measurement unit in its body to provide orientation. An onboard NVIDIA Jetson Xavier NX computer runs the control policy at 50 Hz.

\textbf{Robot Arm}: A Unitree Z1 arm is mounted on the B1's base. This \SI{6}{} degree-of-freedom robot arm weighs \SI{5.3}{\kilo\gram} and has a maximum reach of \SI{0.74}{\meter}.

\textbf{Teleoperation Joystick}: To control the robot and arm, we use the Oculus Meta Quest 2 with its two controllers or the Apple Vision Pro headset  (Figure \ref{fig:system_diagram}). When using the Meta Quest 2, the motion of the right controller is mapped to position commands for the end effector. The robot's body motion, force application, and position commands at the end effector are controlled by the thumb joysticks and buttons on both controllers. The last two arm joints, responsible for controlling the gripper's roll orientation and opening angle, are directly commanded via joystick commands, as they do not influence the force or position of the end effector. 

\textbf{Simulator}: We use Isaac Gym \cite{isaacGym} for policy training.

\section{Method}
\label{sec:method}
For many tasks, the robot must switch between following position commands (e.g., walking to grasp the door handle) and applying forces (e.g., rotating the door handle to open the door) (Section \ref{sec:eepos_loc_task}). We learn a single policy that simultaneously learns locomotion and whole-body regulation of the end effector position and force. The learned policy is $a_t = \pi_{\theta}(o^{t-H:t})$ where $o^t$ is the agent's observation at time step $t$ and $H=30$ is the length of the observation history window.
Our primary contribution is the ability to track forces (Section \ref{sec:force_control}) which enables compliant and force-controlled tasks. 

\subsection{Action and Observation Space}
\label{sec:actobs}
The action space is seventeen-dimensional ($a_t \in \mathbb{R}^{17}$), controlling position targets for a proportional-derivative controller in each of the robot's joints: thigh, calf, and hip joints of the B1 robot as well as the first five joints of the arm.
The position targets are computed as $\sigma_a a_t + q_{def}$, where $\sigma_a = 0.25$ is a scaling factor, $a_t$ is the policy’s output, and $q_{def}$ denotes the default joint configuration, reflecting the robot's standard standing pose with its arm raised. 

The observation, denoted as $o^t$, consists of  
the gravity vector projected in the body frame $g_{base}^t \in \mathbb{R}^{3}$, the feet clock timings $\theta_{feet}^t \in \mathbb{R}^{4}$ \cite{margolis2022walktheseways}, the joint positions, $q^t \in \mathbb{R}^{17}$, the joint velocities, $\Dot{q}^t \in \mathbb{R}^{17}$, 
and the previous actions $a^{t-1} \in \mathbb{R}^{17}$:
\begin{equation}
    o^t = [g_{base}^t, \theta_{feet}^t, q^t, \Dot{q}^t, a^{t-1}] \in \mathbb{R}^{58} 
\end{equation}

The observation history $o^{[t-H,.., t-1, t]}$ ($H=30$) is concatenated to the history of task-associated commands $c_{cmd}^{[t-H,.., t-1, t]}$ as input to the policy.

\begin{figure}[t!]
\centering

\includegraphics[width=0.78\linewidth]{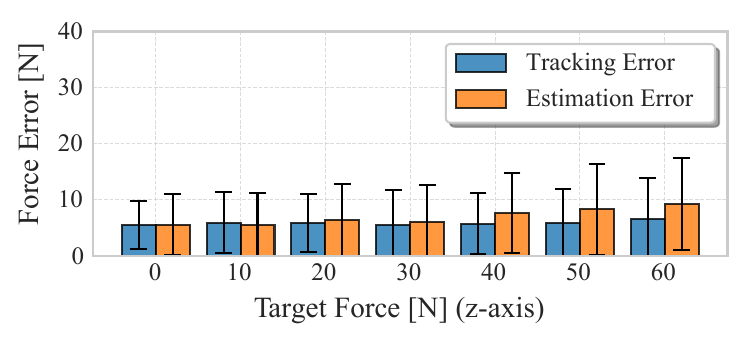}
\raisebox{0.3\height}{\includegraphics[width=0.2\linewidth]{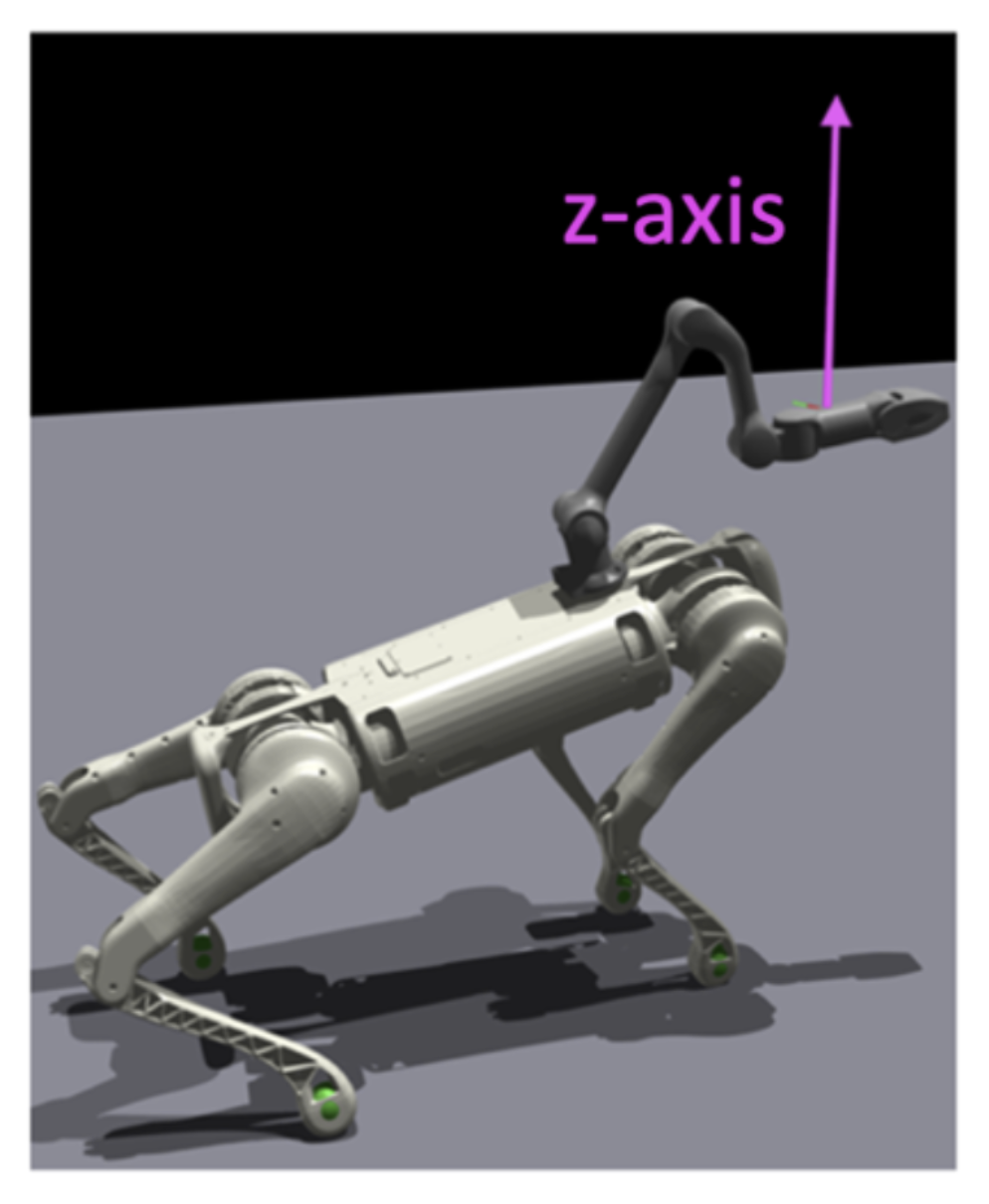}}

\vspace{-0.2cm}
\caption{\textbf{Simulated force control evaluation.} Average force tracking and estimation errors for target forces applied 
at 1000 setpoints sampled across the full training workspace. The bars represent the standard deviation. The accompanying illustration (right) shows how the robot applies force at the gripper along the z-axis.}
\label{fig:force_z_sim}
\end{figure}

\subsection{Task Definition}
The policy input includes four commands :
\begin{itemize}
    \item The desired force vector at the end effector $F^\textrm{cmd} \in \mathbb{R}^3$ represented in Cartesian coordinates in the world frame in the range of \SI{-70}{\newton} and \SI{70}{\newton} for each axis.
    \item The desired end effector position in the body frame is expressed in spherical coordinates: $p^\textrm{cmd}_{ee}=(r, \theta, \phi)^{cmd} \in \mathbb{R}^3$, where $r^{\text{cmd}} \in$ [\SI{0.3}{}, \SI{0.9}{\meter}], $\theta^{\text{cmd}} \in$ [-0.4$\pi$ \SI{}{}, 0.4$\pi$ \SI{}{\radian}], $\phi^{\text{cmd}} \in$  [-0.6$\pi$ \SI{}{}, 0.6$\pi$ \SI{}{\radian}].
    \item The desired linear velocity in the body frame x- and y- axes and the yaw angular velocity: $v_x \in [-1, \SI{1}{\meter/\second}]$, $v_y \in [-1, \SI{1}{\meter/\second}]$, $\omega_z \in [-1, \SI{1}{\radian/\second}]$. To make teleoperation easier, when the magnitude of the commanded base velocity falls below $0.1$, the desired contact schedule for each foot is set to zero, ensuring all feet remain grounded. 
    This prevents the robot from stepping in place, easing teleoperation. 
    \item The binary input $C^f$ indicates whether the robot is in position or force tracking mode.
\end{itemize}

During the force control task ($\mathbf{C}^f = 1$), the position commands $p^\textrm{cmd}_{ee}$ and position tracking rewards $\mathbf{r}_x^g$ are set to zero. Conversely, during the end effector positioning task ($\mathbf{C}^f = 0$), the force commands $F^\textrm{cmd}$ and force tracking rewards $\mathbf{r}_f^g$ are set to zero. During training, this input is resampled randomly twice within each episode, so the training alternates between force and position tracking.

\subsection{End Effector Force Task} \label{sec:force_control}
\label{sec:force_controller}

To learn force control, we need to create a training setup that models how the robot's actions result in applied forces when it makes contact with the environment. If the training consisted of reaching through free space, as is typical in prior works, the robot's gripper would never 
experience any external forces besides self-collisions. It would, therefore, accelerate rapidly to the workspace limit when applying a constant force. 
Such acceleration can be prevented if the robot experiences external forces. To make the robot experience varied forces that emulate different kinds of contact, we simulate external force applied to the robot by implementing a soft contact model~\cite{nagurka2004mass}. 
The simulated external force field, $F_e(\mathbf{x}_g^t) \in \mathbb{R}^3$, applies force to the gripper as a function of its displacement from a position setpoint according to a spring-damper model.
The force is modulated using a proportional-derivative control scheme on the position and velocity difference between the gripper position, $\mathbf{x}_g^t \in \mathbb{R}^3$ and the setpoint $\mathbf{x}_s\in \mathbb{R}^3$:

\begin{equation}
    F_e^t = K_p^\textrm{ext} (\mathbf{x}_s - \mathbf{x}_g^t) + K_d^\textrm{ext} (\Dot{\mathbf{x}_s} - \Dot{\mathbf{x}_g}^t)
\end{equation}

\begin{figure*}[ht!]
    \vspace{0.3cm}
    \centering
    \includegraphics[clip,trim=0cm 0.3cm 0cm 0.2cm,width=0.75\linewidth]{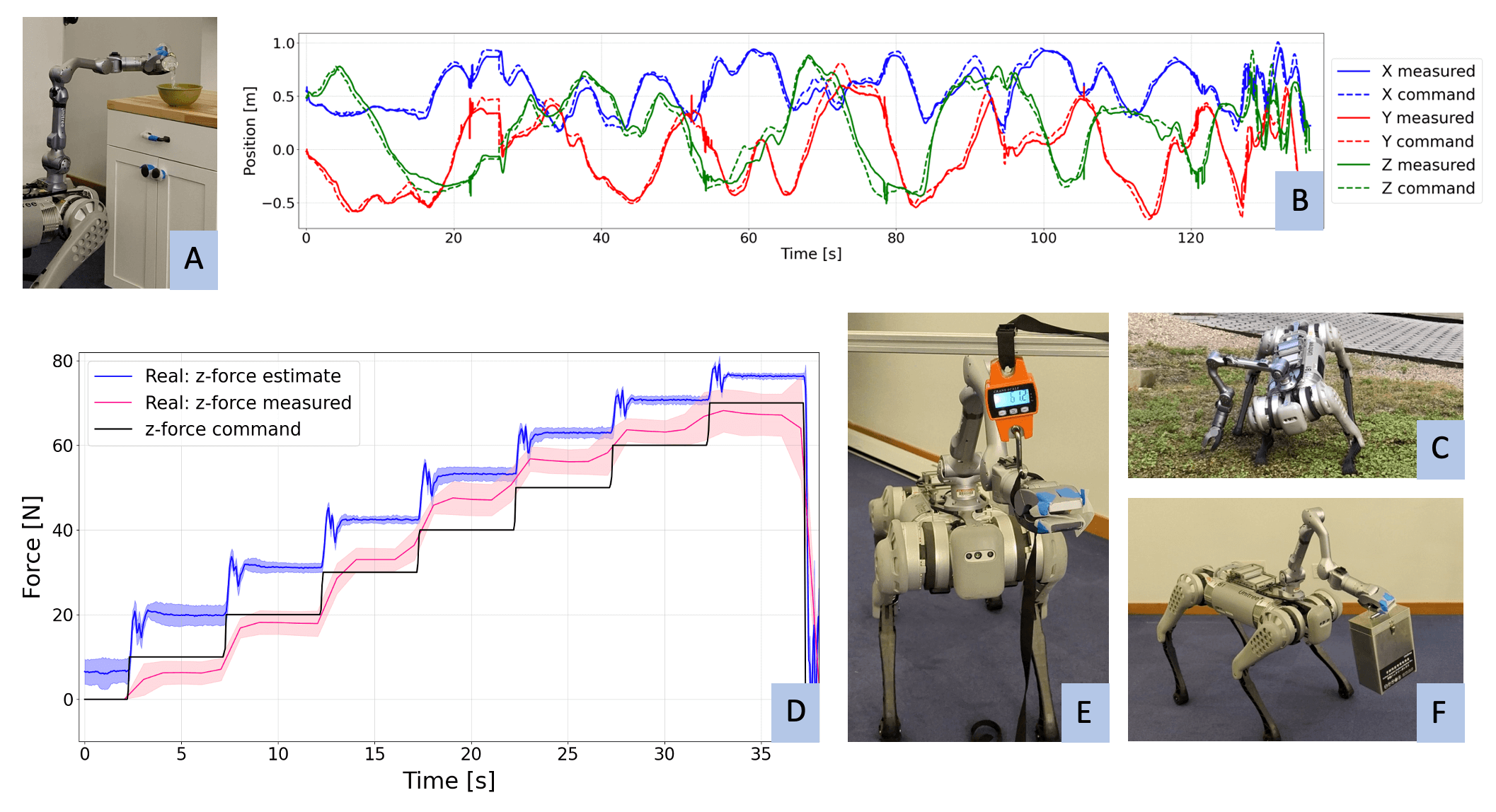}\hfill
    \caption{\textbf{Real-world deployment.} We deploy our policy in a variety of real-world scenarios: (A) water pouring in a kitchenette; (B) tracking a long end effector trajectory in position control mode with performance verified by motion capture; (C) outdoor whole-body reaching; (D, E) quantitative measurement of the force application control by pulling down on a dynamometer; (F) lifting a \SI{4}{\kilo\gram} box while maintaining compliance in the gripper, showcasing gravity compensation. In subfigure (D), each line represents the mean across five force setpoints, with the shaded area indicating the standard deviation.
    }\label{fig:sim_to_real}
\vspace{-0.4cm}
\end{figure*}

The chosen gains, $K_p^\textrm{ext} = 700$, $K_d^\textrm{ext} = 6$, were tuned by inspecting the behavior of the simulator to ensure that the applied force is large enough to bring the gripper to the setpoint but small enough to avoid extreme torques or oscillations. The setpoint is defined with respect to the robot's body frame and moves with the robot during locomotion.

If we initialize the position setpoint $\mathbf{x}_s$ randomly, it may be unreachable or in collision with the robot's body. To avoid this, we instead initialize the position setpoint at the gripper's initial position when alternating between the position tracking task and the force tracking task. As the robot learns to reach end effector positions across a wide workspace, it simultaneously acquires proficiency in exerting force across that workspace.

We repeatedly sample the force target $F^\textrm{cmd}$ and the force application duration $t_F$, which ranges between two and four seconds. The force command undergoes linear interpolation from zero to the sampled values in $t_F$ seconds. Then, they maintain these values for a duration of 1.5 seconds, after which they are linearly interpolated back to zero in $t_F$ seconds. To ensure that fully compliant behavior is practiced frequently, each time a force command is sampled, there is a 20\% chance of being a zero force target.

\subsection{End Effector Positioning Task} 
\label{sec:eepos_loc_task}
The end effector position control mode is a stepping stone towards our primary goal of forceful manipulation. This mode  (i) supports walking to objects and grasping them (Section \ref{sec:posresults}), and 
(ii) generates diverse stable and collision-free initialization poses for the force mode (\ref{sec:force_control}). This mode should have a wide workspace and intuitive teleoperation interface for reaching any point in the environment. This control mode reimplements elements from prior work \cite{fu2023deep} with some adaptation to suit our robot's specifications and facilitate alternation with force training. We adopt the command parameterization from \cite{fu2023deep}, where the end effector commands are defined within a frame centered around the body and remain invariant to body height, roll, or pitch changes. Each time a new position is sampled, we linearly interpolate the position target from the previous position command to the new one for four seconds.

\begin{table}[t!]
    \centering
    \vspace{0.3cm}
    \bgroup
    \def\arraystretch{1.2}
    \tiny
    \begin{tabular}{llr}
    \toprule
    \textit{Term} & \textit{Equation} & \textit{Weight} \\ [0.5ex]
    \hline
    \multicolumn{3}{|c|}{End Effector Position Control (\ref{sec:eepos_loc_task})} \\
    \hline
    $\mathbf{r}_x^g$: gripper position & $\exp\{-|p_{ee} - p_{ee}^{\textrm{cmd}}|/0.5\}$ &  \tabnode{$5.0*\neg C^f$} \\[0.5ex]

    \hline
    \multicolumn{3}{|c|}{End Effector Force Control (\ref{sec:force_control})} \\
    \hline
     $\mathbf{r}_f^g$: gripper force  &  \( \exp\{{-{|F-F^{\text{cmd}}|} / {20}}\}\) & \tabnode{$5.0*C^f$} \\ [0.5ex]

    \hline
    \multicolumn{3}{|c|}{Safety and Smoothness ($\mathbf{r}_l$)} \\
    \hline
    collision penalty & \( \mathds{1}_{\text{collision}}\) & $-5.0$ \\ [0.5ex]
    arm joint limit & \( \mathds{1}_{q_{a} > 0.9*q_{a}^{max} || q_a < 0.9*q_a^{min}}\) & $-3.0$ \\ [0.8ex]
    leg joint limit & \( \mathds{1}_{q_{l} > 0.9*q_{l}^{max} || q_l < 0.9*q_l^{min}}\) & $-1.0$ \\ [0.8ex]  
    joint velocities & \(|\dot{q}|^2\) & \num{-8e-4} \\ [0.5ex]
    joint acceleration & \(|\ddot{q}|^2\) & \num{-3e-7} \\ [0.5ex]

    action smoothing & \(|a_{t-1} - a_{t}|\) & \num{-0.05} \\ [0.5ex]
    -- & \(|a_{t-2} - 2a_{t-1} + a_{t}|^2\) & \num{-0.02} \\ [0.5ex]
    arm torque limit &  \(\mathds{1}_{\mathbf{\tau}_{a} > 0.8*|\mathbf{\tau}_{a}^{max}|} \) & \num{-0.0015}\\ [0.5ex]
     \hline
    \multicolumn{3}{|c|}{Locomotion ($\mathbf{r}_{v}^b$)}  \\
    \hline
    body velocity & \( \exp\{{-{|v_{b}-v^\textrm{cmd}_{b}|} / {0.25}}\}\)& \tabnode{$1.0$} \\ [0.5ex]
    swing phase & \( \sum_{\text{foot}} [1-C_{i}^{cmd}(t)] \exp\{{-{|f^{\text{foot}}|^2} / {4}}\} \) & $0.9$ \\ [0.5ex]
    stance phase & \( \sum_{\text{foot}} [C_{i}^{cmd}(t)] \exp\{{-{|v^{\text{foot}}_{xy}|^2} / {4}}\} \) & $4.0$ \\ [0.5ex]

    \bottomrule
    \end{tabular}
    \egroup
    \vspace{0.1cm}
    \caption{Reward terms for learning the whole-body policy.}
    \vspace{-0.5cm}
    \label{tbl:rewards}
\end{table}

\subsection{Reward Function, Policy Architecture, and Optimization}
\label{sec:policy_arch}

The reward is ${(\mathbf{r}_v^b + \mathbf{r}_x^g + \mathbf{r}_f^g)e^{\mathbf{r}_l }}$ with separate terms for the locomotion task ($\mathbf{r}_v^b$), gripper force control task ($\mathbf{r}_f^g$), gripper position control task ($\mathbf{r}_x^g$), and safety and smoothness criteria ($\mathbf{r}_l$) (Table \ref{tbl:rewards}). To encourage a smooth gait, the locomotion task includes a contact schedule pattern for trotting~\cite{margolis2022walktheseways}. The exponential form from~\cite{ji2022concurrent} ensures the reward will remain positive, disincentivizing early termination.

The policy itself includes the actor and state estimation modules, while the entire system comprises three modules: the actor network, the critic network, and the estimation module.
Following a common optimization technique in sim-to-real learning~\cite{chen2020learning, ji2022concurrent}, we define a privileged state consisting of quantities that may aid learning in simulation but are not available from the real-world sensors. Our privileged state includes the robot's body velocity, the gripper position in the body frame, and the external force on the gripper.
First, the state estimation module predicts an estimate $\hat{e}_t$ approximating the privileged state $e_t$ from the observation history. This style of concurrent state estimation can improve optimization performance \cite{ji2022concurrent}. Then, the actor-network inputs the observation history and the state estimate $\hat{e}_t$ and outputs the action. Separately, the critic network inputs the observation history and the true privileged state $e_t$. The estimator, actor, and critic network are multilayer perceptions with \texttt{elu} nonlinearities and hidden layer dimensions $[256, 128]$, $[512, 256, 128]$, and $[512, 256, 128]$, respectively. The state estimator is simultaneously trained with a supervised loss while the actor and critic networks are optimized using Proximal Policy Optimization \cite{schulman2017proximal} with 4096 parallel environments. The maximum episode length is 20 seconds which corresponds to 1000 timesteps.
To speed up training, episodes are terminated if the agent reaches a failure state. 
The robot is deemed to fail if the gripper is in collision or if the body height falls below \SI{0.3}{\meter}.

\section{Results}
\label{sec:results}


\subsection{Force Control Policy Performance}

\label{sec:large_forces}

When lifting or pulling objects, an effective controller should realize a large applied force without undesired transients or inefficient postures that would result in the motor exceeding its safety limits. To evaluate this characteristic, the average force tracking and estimation errors for target forces applied at 1000 setpoints sampled across the full training workspace are reported in Figure \ref{fig:force_z_sim}. The tracking error represents the difference between the commanded and actual force, while the estimation error denotes the difference between the estimated and actual force. For z-axis force control, the mean absolute tracking error is around \SI{5}{\newton} for low force targets and remains below \SI{10}{\newton} across the entire training range. The estimation error is generally equal to or greater than the tracking error, suggesting that some errors may result from a lack of observability in the policy input.

To evaluate the force control performance on the real platform, we attach our robot's end effector to a dynamometer and gradually increase the downward force command through the entire training range (\SI{0}{}-\SI{70}{\newton}). The dynamometer is used to record the applied force across five trials with the gripper in high, low, left, right, and middle positions (Figure \ref{fig:sim_to_real}-E). 
We found that the tracking error across these five trials was within the range of \SI{5}{}-\SI{10}{\newton} which is similar to the values observed in simulation. The estimated force tends to overshoot, suggesting a moderate sim-to-real gap. Despite significant discrepancies in force estimation, the force tracking performance remains notably proficient, suggesting that the policy somewhat disregards the estimated force feedback.

To evaluate whether our policy can coordinate the body with the legs to increase the applied force, we initialized the arm directly in front of the robot and recorded the highest pulling force we can achieve (Figure \ref{fig:main}). The observed force of \SI{90}{\newton} is greater than the arm's rated payload of \SI{36}{\newton}. We also measured the force application across the reaching workspace in simulation. The policy can track forces across a large portion of the expanded workspace, although the error is higher near the edge of the workspace (see supplement).

\subsection{Impedance Control}
Rather than commanding the desired force directly, a common use of force control is to implement an impedance controller, where the end effector tracks a target position while complying to external forces in the manner of a spring-mass-damper system~\cite{hogan1984impedance, hogan1985impedance}. We implement an impedance controller using our force control policy with no additional training by introducing a feedback loop on the policy inputs and outputs: $\textbf{F}_{cmd} = -K_p^\textrm{imp}(\textbf{x} - \textbf{x}_{des}) - K_d^\textrm{imp} \dot{\textbf{x}}$ where $\textbf{x}_{des}$ is the target position in the body frame and $\textbf{x}$, $\dot{\textbf{x}}$ are the current gripper position and velocity in the body frame. To obtain the current gripper position and velocity, we read them from the output $\hat{e}_t$ of the learned state estimation module (Section \ref{sec:policy_arch}).  In the supplementary video, we demonstrate the teleoperation of the robot under impedance control to grasp a chair low to the ground and drag it while walking.

\subsection{Compliance and Kinesthetic Demonstration}

\label{sec:applications}

\subsubsection{Compliant end effector state}

When the controller \ref{sec:force_controller} is commanded to track zero force, this corresponds to a fully compliant mode where the posture of the body and arm coordinate to drive the gripper force application to zero in all three axes. When released, the gripper remains suspended, with the system exhibiting gravity compensation. Figure \ref{fig:main} and the supplementary video show an operator manipulating the system to reach various points. Our result supports that compliant behavior is possible using the standard reinforcement learning architecture. Realizing compliant behavior in learned motor policies will likely improve the safety of robots around humans and their robustness against unexpected disturbances. For example, when the quadruped walks in force control mode and bumps its gripper into a wall or obstacle, we observe that the gripper complies and moves out of the way, allowing the robot to continue walking as desired. Compliance also facilitates data collection for kinesthetic teaching, in which a human operator directly manipulates a robot's limbs to demonstrate a task without writing code or learning to use a teleoperation interface. As a proof of concept, we kinesthetically manipulated the robot to insert a drill into its charging station.

\subsubsection{Compliant manipulation of heavy objects}

By modulating the force application command in the z-axis, the compliant mode can be extended to the scenario where the robot is lifting an object with its arm (Figure \ref{fig:sim_to_real}-F). We tested the compliant mode with payloads of \SI{0}{\kilo\gram}, \SI{2}{\kilo\gram}, \SI{4}{\kilo\gram}, \SI{6}{\kilo\gram}. With a zero force command along the z-axis, nonzero payloads cause the gripper to sink to the ground, which is expected since it should not apply a resistive force. In this case, it takes substantial force for the human to lift the gripper with the object in grasp because the gripper will not apply any lifting force to the grasped object. Next, we increased the vertical force command to match the payload weight. For payloads up to \SI{4}{\kilo\gram}, this restored gravity compensation against the payload and compliance in all axes. With the highest payload of \SI{6}{\kilo\gram}, the gripper is less compliant and drifts to the center of the robot when released to the side.

\subsection{Walking, Reaching, and Grasping}
\label{sec:posresults}

\begin{figure}[t!]
\centering
\begin{subfigure}{0.54\linewidth}
\centering
\includegraphics[width=1.0\textwidth]{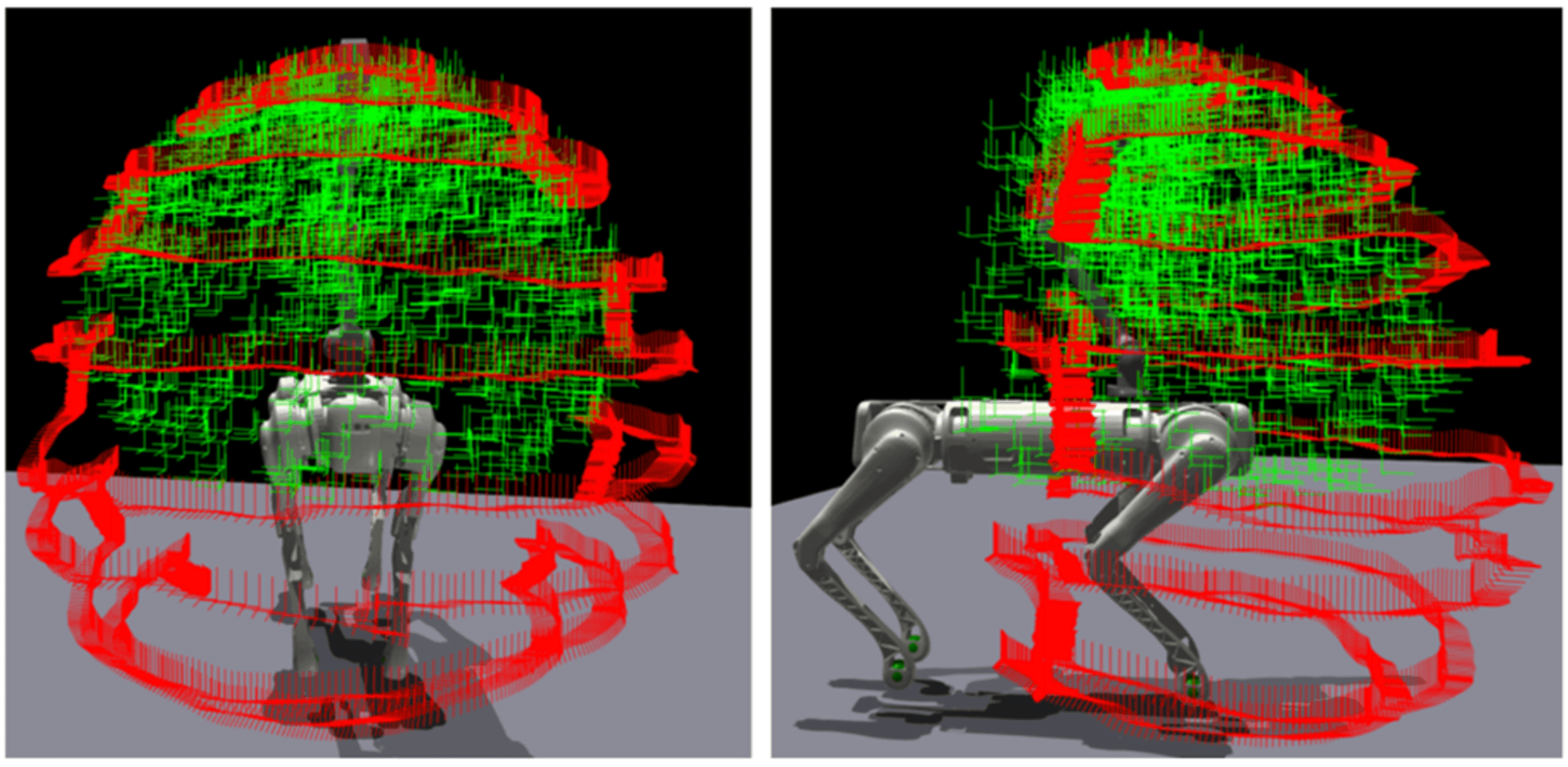}
\caption{Reaching workspace}
\label{fig:whole_hull}
\vspace{0.5em}
\scriptsize
\def\arraystretch{1.2}
\begin{tabular}{|l|c|}
\hline
\textbf{Controller type} & \textbf{Workspace [m\textsuperscript{3}]}  \\ \hline
Arm & 0.81 \\
Whole body & 1.29 \\ \hline
\end{tabular}
\caption{Workspace convex hull volumes}
\label{table:vol}
\end{subfigure}
\hfill
\begin{subfigure}{0.43\linewidth}
\centering
\includegraphics[width=0.75\textwidth]{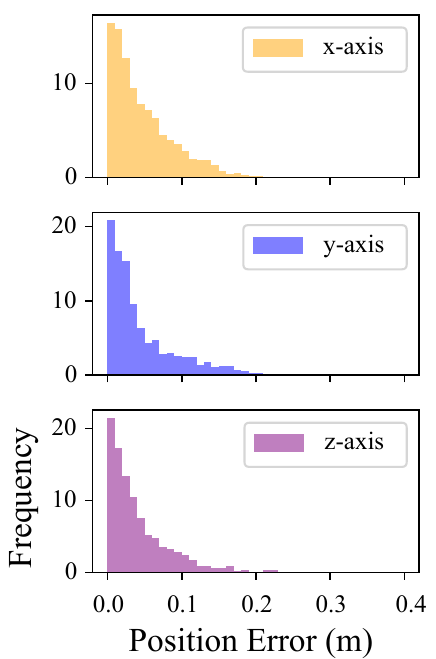}
\caption{Position error histogram}
\label{fig:hist}
\end{subfigure}

\caption{\textbf{Expanded workspace.} (A-B) The whole-body policy in position control mode (red) enlarges the fixed arm workspace (green) by 59\%. (C) The position error distribution shows the tracking accuracy while standing in place, evaluated across 1000 end effector positions sampled from the training distribution.}
\label{fig:wbc_workspace}
\end{figure}

The end effector position control mode helps to establish an initial grasp or contact with the force application target. To manipulate objects that are higher or lower than the robot's body, we would like to achieve a large manipulation workspace through coordination of the body and arm. To measure the workspace of the position reaching, we send a sequence of commands at the limit of the training distribution and record the actual gripper positions achieved. As a baseline, we measure the workspace of the arm while the base is fixed in place by randomly sampling 3000 arm configurations and recording the resulting gripper positions. To characterize the workspace in each scenario, we fit a convex hull to the reached points and compute its volume. The whole-body policy enlarges the workspace volume by \SI{59}{\percent}, demonstrating the significant benefit of whole-body coordination (Figure \ref{fig:wbc_workspace}).
We measure the distribution of error between the end effector position and the target position across a set of $1000$ end effector position commands, uniformly sampled from the training distribution. The mean error is \SI{4.6}{\centi\meter} in the x-axis, \SI{4.8}{\centi\meter} in the y-axis, \SI{5.5}{\centi\meter} in the z-axis.

The end effector tracking performance is evaluated on hardware by teleoperating the arm across a variety of commands within the training distribution. We record the trajectory of commands as well as the trajectory of end effector positions captured via a motion-tracking system. The path, expressed in Cartesian coordinates, can be seen in Figure \ref{fig:sim_to_real}-B. As shown, the commanded and actual end effector positions align closely. For this trajectory, the average positional errors on the $x$, $y$, and $z$ axes are 4.42 cm, 5.37 cm, and 6.86 cm, respectively. These values being similar to
the ones observed in simulation, the disparity between simulation and real-world performance in the position mode is minimal. In a series of qualitative experiments, we successfully teleoperated the end effector positioning mode for door opening (Figure \ref{fig:main}) and water pouring (Figure \ref{fig:sim_to_real}-A).

\section{Discussion}

We have demonstrated that learned whole-body manipulation policies can acquire a degree of compliance and perform force control at the end effector using only the minimal sensor configuration of joint encoders and body IMU. The force tracking performance is sufficient for some force-controlled tasks, including collecting demonstrations for kinesthetic teaching with a weight and whole-body pulling. In quantitative experiments, we demonstrate good accuracy of a learned force estimator and the force application command tracking. We also characterize the sim-to-real gap in our system. Because the policy can walk, grasp, and apply forces, our work provides a teleoperation framework suitable for teleoperation and data collection for kinesthetic teaching of forceful loco-manipulation tasks. In the future, it will be promising to explore imitation learning pipelines where the robot learns to accomplish compliant and forceful behaviors autonomously from demonstrations. 

Our approach differs from some successful classical methods by realizing force control through a low-frequency neural network policy and without any force-torque sensor at the end effector. Although we showed that our policy can realize desired forces and positions, the achieved tracking accuracy of around \SI{5}{\centi\meter} for position and \SI{5}{\newton} for force commands may be insufficient for highly precise tasks. It is unknown whether this performance represents a limit of the hardware, optimization method, or choice of policy architecture. Future work could explore modifications to guide the policy to a more precise solution.

\section*{Acknowledgement}
\footnotesize{
We thank the members of the Improbable AI lab for helpful discussions and feedback. We are grateful to MIT Supercloud and the Lincoln Laboratory Supercomputing Center for providing HPC resources. This research was partly supported by Hyundai Motor Company, the MIT-IBM Watson AI Lab, and the National Science Foundation under Cooperative Agreement PHY-2019786 (The NSF AI Institute for Artificial Intelligence and Fundamental Interactions, http://iaifi.org/). This research was also sponsored by the United States Air Force Research Laboratory and the United States Air Force Artificial Intelligence Accelerator and was accomplished under Cooperative Agreement Number FA8750-19-2-1000. Research was sponsored by the Army Research Office and was accomplished under Grant Number W911NF-21-1-0328. The views and conclusions contained in this document are those of the authors and should not be interpreted as representing the official policies, either expressed or implied, of the United States Air Force or the U.S. Government. The U.S. Government is authorized to reproduce and distribute reprints for Government purposes, notwithstanding any copyright notation herein.
}

\section*{Author Contributions}
\small{
\begin{itemize}
\item \textbf{Tifanny Portela} contributed to ideation, implementation of the entire system, experimental evaluation, and writing.
\item \textbf{Gabriel B. Margolis} contributed to ideation, implementation of some parts of the system, and writing.
\item \textbf{Yandong Ji} contributed to ideation and initial infrastructure.
\item \textbf{Pulkit Agrawal} advised the project and contributed to its
development, experimental design, and writing.
\end{itemize}
}

\bibliographystyle{IEEEtran}
\bibliography{IEEEabrv,IEEEfull}{}

\end{document}